\pdfoutput=1
\documentclass[11pt]{article}

\usepackage{EMNLP2023}

\usepackage{times}
\usepackage{latexsym}
\usepackage{booktabs}
\usepackage{multirow}
\usepackage{graphicx}
\usepackage{makecell}
\usepackage{tabularx}
\usepackage{color}

\usepackage[T1]{fontenc}
\usepackage[utf8]{inputenc}

\usepackage{inconsolata}
\usepackage{caption}
\usepackage{amsfonts}

\usepackage{microtype}

\title{GPT4Graph: Can Large Language Models Understand Graph Structured Data? An Empirical Evaluation and Benchmarking}

\author{Jiayan Guo$^{1}$\footnote{}, Lun Du$^{2}$\footnote{Corresponding Author}, Hengyu Liu$^{3}$, Mengyu Zhou$^{2}$, Xinyi He$^{4}$, Shi Han$^{2}$ \\
  $^{1}$School of Intelligence Science and Technology, Peking University; \\ $^{2}$Microsoft; $^{3}$University of Technology Sydney; $^{4}$Xi’an Jiaotong University \\
  \texttt{guojiayan@pku.edu.cn}, \texttt{\{lun.du,mezhou,shihan\}@microsoft.com}, \\ \texttt{hengyu.liu@uts.edu.au},\texttt{hxyhxy@stu.xjtu.edu.cn}
  }

\begin{document}
\maketitle
\begingroup\def\thefootnote{*}\footnotetext{Work done during the internship at MSRA.}
\endgroup
\begingroup\def\thefootnote{$\dagger$}\footnotetext{Corresponding Author.}
\endgroup
\begin{abstract}
  Large language models~(LLM) like ChatGPT have become indispensable to artificial general intelligence~(AGI), demonstrating excellent performance in various natural language processing tasks. Graph data is ubiquitous and an essential part of AGI. The training corpus of large language models often includes some algorithmic components, which allows them to achieve certain effects on some graph data-related problems. However, there is still little research on their performance on a broader range of graph-structured data. In this paper, we conduct an %\origin{investigation}
  empirical study to assess the proficiency of LLMs in comprehending graph data, employing a diverse range of structural and semantic-related tasks that evaluate the LLMs' capabilities in graph understanding. Through our study, we uncover current limitations and future directions of LLMs in comprehending graph and performing associated reasoning tasks.
    % \lun{Comment: the second half of the sentence (after ``but also'') is not very informative.}.
\end{abstract}

\section{Introduction}

Large Language Models (LLMs) have demonstrated significant capability across a diverse array of human-centric tasks. These tasks range from answering questions to performing semantic analysis and identifying named entities~\cite{zhao2023survey}. Despite the considerable strides that have been made, the capacity of LLMs to decipher and manage structured knowledge, especially in the form of graph-structured data, remains an area ripe for exploration. Understanding graph-structured data is vital, given its pervasive presence and integral role in a multitude of applications such as social network analysis, drug discovery, recommender systems, and spatio-temporal prediction. Understanding graph data is crucial for AGI.

Tasks based on graph data can be broadly classified into two categories based on their goals. The first category includes structure understanding tasks like identifying significant nodes, calculating centrality metrics~\cite{okamoto2008ranking,zhang2017degree,brandes2001faster,barthelemy2004betweenness,newman2005measure}, and determining diameters~\cite{chung1994upper}. The second category encompasses semantic understanding tasks, such as knowledge graph question answering~(can be abstracted as knowledge graph )~\cite{huang2019knowledge,zhang2018variational}, node classification~\cite{bhagat2011node,rong2019dropedge} and graph classification~\cite{errica2019fair}, etc. These tasks have distinct requirements and challenges.

Previous researches have investigated the use of LLMs for structural understanding~\cite{sui2023evaluating,jiang2023structgpt,gong2020tablegpt,liu-etal-2022-plog}, but the emphasis has been predominantly on tables, which rely heavily on structured tabular data. Graphs, on the other hand, introduce additional dimensions of complexity. Comprised of nodes that represent entities or concepts, and edges that express relationships between these entities, graphs necessitate a more sophisticated level of comprehension from LLMs. Understanding graph structued data with LLM remains challenges. First of all, graph data can not be directly handled by LLM, as graph data are unorganized and complex. Secondly, there is a wide range of graph-related tasks, designing efficient input format for different tasks and effective prompt techniques is essential while rarely explored.

In this paper, our goal is to setup a comprehensive comparison to show the ability of LLM in understanding graph structured data. To achieve this goal, we first bridge the existing gap between Large Language Models (LLMs) by proposing a novel framework that integrates LLMs and graph-structured data, intending to enhance their synergistic ability across a wide range of graph mining tasks. Based on the framework, we establish a benchmark across ten common scenarios to assess language models' capability in handling graph-related tasks. In addition, we experiment with various prompting methods, including both handcrafted and self-generated prompts, to demonstrate their effectiveness in boosting performance in both zero-shot and few-shot settings. Our findings reveal that while LLMs have demonstrated some capability in handling graph-structured data, there remains a substantial need for further development to achieve a performance level comparable to specialized graph-oriented models. In summary, our contribution can be summarized by:

\begin{itemize}
\item We introduce a new framework that combines Large Language Models (LLMs) and graph-structured data. This setup uses the language understanding skills of LLMs and graph description language with promt engineering to improve how they work together in different situations.
\item We develope a wide-ranging set of tasks, across ten common scenarios, to check how well LLMs can handle tasks involving graph data. This set of taks provides a consistent way to check how good language models are at dealing with complex graph data.
\item Our empirical results show that, while LLMs are getting better at handling graph data, they still have a lot of improving to do if they are to catch up with models that are specifically designed to work with graphs.
\end{itemize}

\section{Preliminary}

\subsection{Graph Mining Tasks}

Graph mining tasks refer to the process of extracting valuable and actionable insights from graph-structured data. Graphs are mathematical structures that represent relationships between entities, where nodes represent entities and edges represent the connections or interactions between them. Graph mining involves analyzing these graphs to discover patterns, relationships, communities, and other useful information. Some graph mining tasks include node classification, link prediction, and community detection. These tasks are crucial in various domains, including social network analysis~\cite{wasserman1994social}, bioinformatics~\cite{baxevanis2020bioinformatics}, recommendation systems~\cite{isinkaye2015recommendation}, fraud detection~\cite{bolton2002statistical}, and knowledge graphs~\cite{ji2021survey}.

\subsection{Graph Description Language}

A graph description language is a formal language or notation used to define or represent graph-structured data. It provides a standardized syntax and semantics for describing the elements and relationships within a graph. Graph description languages enable the creation, manipulation, and interpretation of graphs in a consistent and machine-readable manner. These languages provide a way to define graph structures, specify node and edge properties, and perform queries and operations on graphs. They are essential for working with graph data and enabling interoperability between graph-based systems and tools. For example, graphs can be represented by an edge list or an adjacency list, providing two distinct perspectives on the graph's structure. An edge list defines a graph in terms of its individual connections, whereas an adjacency list describes each node in terms of its neighboring nodes. Along with these basic representations, more sophisticated formats have been developed to convey richer, contextual information about the graph. For instance, the Graph Modelling Language (GML)\cite{himsolt1997gml} and Graph Markup Language (GraphML)\cite{brandes2013graph} provide extensible, language-based frameworks for graph representation.

\begin{figure}
    \centering
    \includegraphics[width=\linewidth]{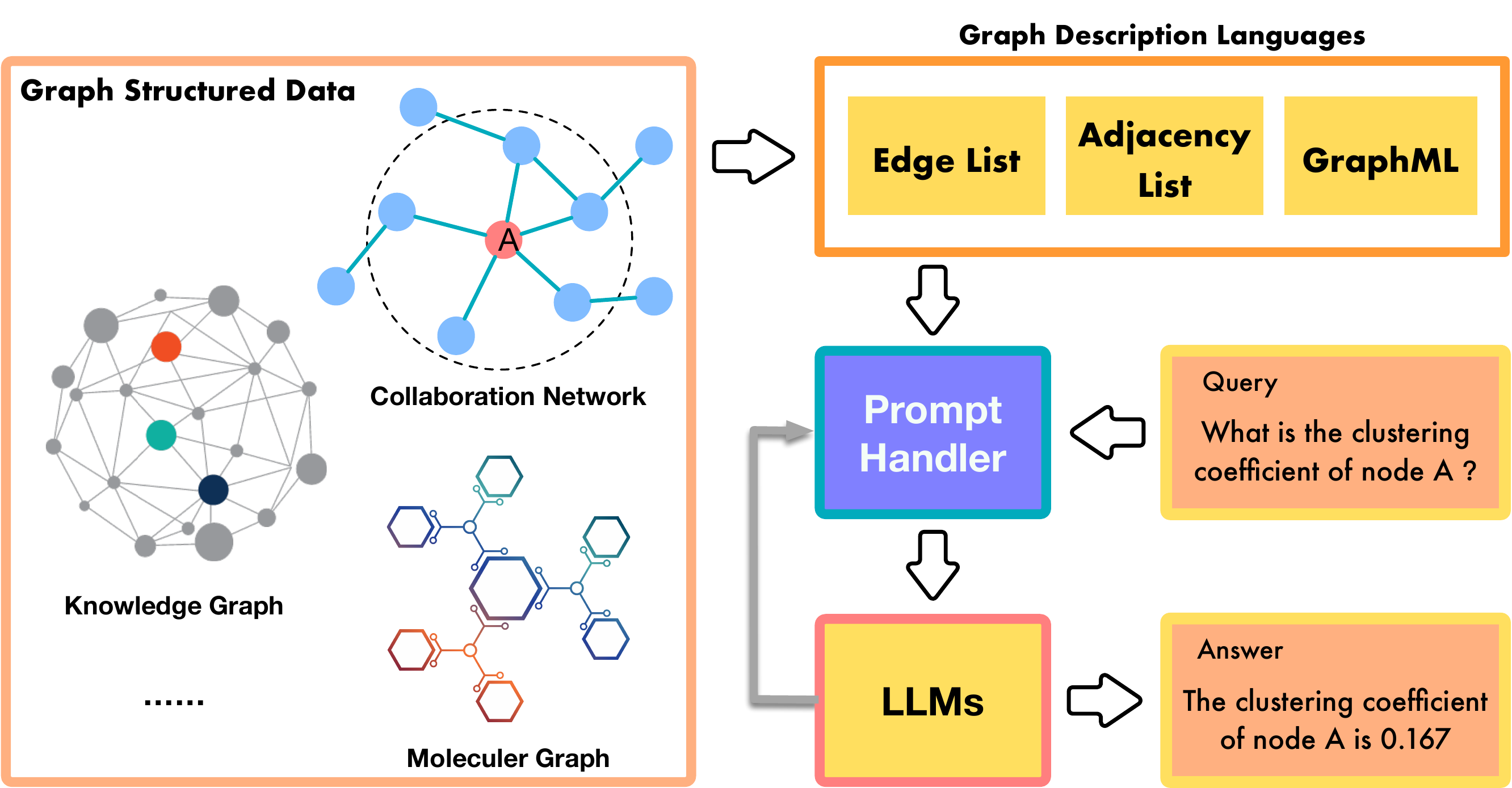}
    \caption{Graph Understanding with LLM Framework. The graph data is first converted to graph description language that can be understand by LLM. Then the prompt handler combines user query and GDL with potential multiple rounds to generate the answer.}
    \label{fig:frametwork}
\end{figure}

\begin{figure*}
    \centering
    \includegraphics[width=\linewidth]{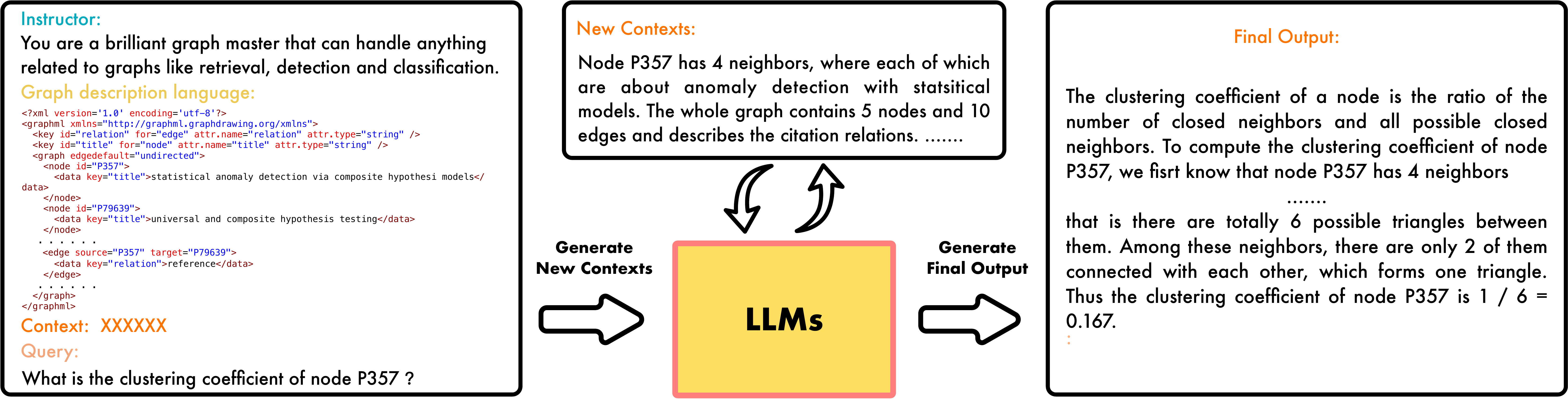}
    \caption{Illustration of Self-prompting. The first request is to ask LLMs to automatically generate the context of the input graph (w/ or w/o respect to the question). We may ask LLMs multiple context related questions. After generating the new context~(such as context summarization and format explanation). the new context is combined with the original input and are send to the LLMs to generate the final output.}
    \label{fig:enter-label}
\end{figure*}

\section{Graph Understanding with LLM Pipeline}

The overall pipline of graph understanding with LLMs is illustrated in Figure~\ref{fig:frametwork}. Where for graph data, we first generate their graph description languages~(GDL), and then use prompt handler to combine the user query and GDL to form the input to the LLMs. The LLMs performs reasoning and generate answers to the user. During the reasoning, the LLMs may generate intermedium output that should be handled by prompt handler to form new input to the LLMs. Here we elaborate the prompt handler to show how to make LLM better understand graph data.

\subsection{Manual Prompting}

Manual prompting for graph-based problems involves utilizing familiar graph representations to prompt a large language model (LLM) for desired outputs. The novelty of this approach lies in the fact that it requires a shift from traditional text-based inputs to graphical representations. These graph formats have been discussed in Section 2.2. By employing these graph formats as input, we can provide more comprehensive and context-rich information about the graph to the LLM. Other manual prompting techniques include adding format explanation to make LLM better understand the format and adding role prompting to make LLM better understand the specific task. Besides, we can also change the input order between question and external input, and adding examples to utilize the in-context learning ability~\cite{weifinetuned} of LLM.

Nonetheless, some recent developed change-of-thoughts promptings~\cite{kojimalarge,yao2023tree} can also be applied to enhance the reasoning ability of LLM for there are many tasks requiring multiple step of reasoning~(e.g clustering coefficient computing). 

\subsection{Self-Prompting}

Sometimes the given graph context contains less useful or redundant information for solving tasks. Thus we need LLM to perform self-prompting to obtain more context or eliminating irrelevant information from the given input. It can be challenging for LLM to generate effective prompts for graph-based tasks, as graphs have complex structures and relationships that need to be accurately captured in the prompt. However, there are several strategies that can be employed for self-prompting in graph-based tasks.

\noindent\textbf{Context Summarization}: LLM can generate a summary of the given graph by extracting key features, such as important nodes, edges, or subgraphs. The generated summary can serve as a prompt for the subsequent graph-related questions or tasks. Besides, based on some important elements like nodes and edges, we can use LLM to summarize their context~(neighborhood) information to form neighborhood aware text features.

\noindent\textbf{Format Explanation}: Sometimes it is hard for a human to give the entire description of the input graph format. To make the LLM gain more context information of the input graph, we can make the LLM to generate format explanation by itself. 

\noindent\textbf{}

% \textbf{Graph exploration}: LLM can simulate the process of exploring the graph by generating a sequence of queries or actions to retrieve information from the graph. These queries can be designed to guide the model in understanding the graph structure and relationships.

% \textbf{Graph completion}: LLM can generate partial 
% graphs and prompt itself to complete the missing parts. This can help the model learn to infer missing information in a graph and develop a better understanding of the graph's overall structure.

By leveraging these self-prompting strategies, LLM can actively engage in the understanding and manipulation of graphs, facilitating graph-based reasoning and learning.

\section{Graph Understanding Benchmark}

\subsection{Structure Understanding Tasks}

\noindent\textbf{Graph Size Detection.} This task evaluates a large language model's (LLM) capability to discern the size of a provided graph. In this context, size refers to the count of nodes and edges present in the graph. The LLM is expected to accurately determine these metrics, even when user-provided designs and accompanying data, such as descriptions, statements, or queries, augment the graph. Despite the inherent challenge this poses for language models, a precise count of nodes and edges is critical, as it enables the LLM to contextualize information accordingly.

\noindent\textbf{Degree Detection.} This task investigates the LLM's aptitude for understanding a node's contextual relevance within a graph. Here, the degree of a node—an indicator of a node's importance and the sparsity of its connections—forms the crux of the task. The LLM must ascertain the number of neighbors for a given node, based on the graph text and any supplementary information. The degree of a node is foundational for various centrality measures such as degree centrality and clustering coefficient, underscoring the task's importance in understanding a node's local structure.

~\

\noindent\textbf{Edge Detection.} Building on degree detection, this task further explores the LLM's understanding of a node's local structure. The model must identify the neighboring nodes of a given node, a skill that is vital for complex graph mining activities like calculating distances and discerning connectivity patterns. Mastery of this task signifies the LLM's comprehension of the fundamental aspects necessary for advanced graph analysis.

~\

\noindent\textbf{Attribute Retrieval.} This task tests the LLM's capacity to retrieve pertinent details about a node, such as the node's attributes, which play a key role in defining its characteristics. For instance, the LLM might need to retrieve a specific attribute such as a paper's title or an author's gender. Success in this task highlights the LLM's ability to comprehend and retrieve essential node-related information.

~\

\noindent\textbf{Diameter Computing.} This task challenges the LLM to calculate the diameter of a graph. The diameter, which is the longest shortest path between any two nodes, offers valuable insights into the graph's overall connectivity and reachability. A successful computation of the diameter showcases the LLM's grasp of the graph's structure and its ability to analyze the graph's overarching characteristics.

~\
\begin{figure}[t]
    \centering
    \includegraphics[width=\linewidth]{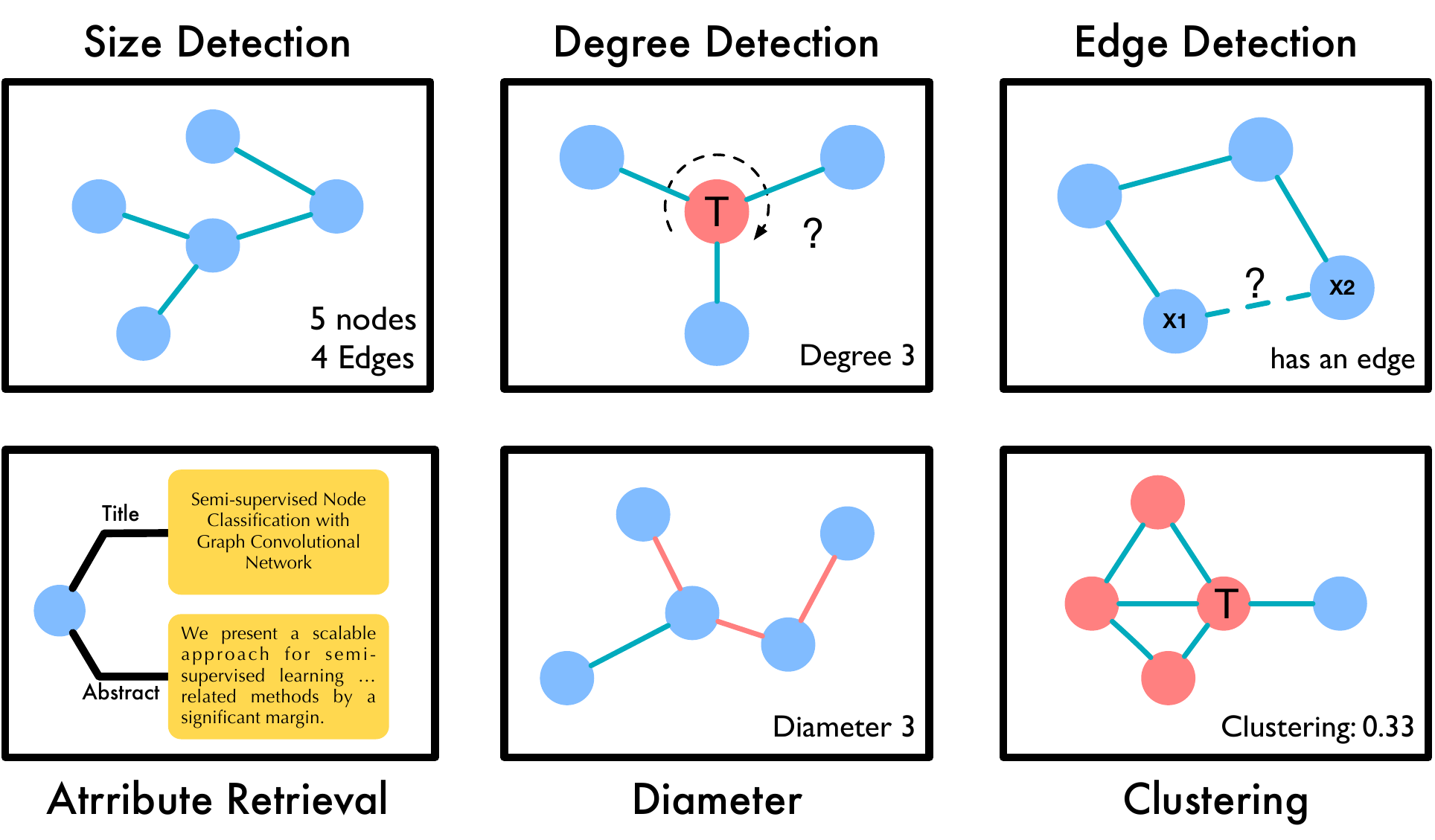}
    \caption{Structure Understanding Tasks}
    \label{fig:sut}
\end{figure}
\noindent\textbf{Clustering Coefficient Computing.} In this task, the LLM needs to compute the clustering coefficient of a graph, a measure that indicates how closely nodes in a graph tend to cluster together. The task thereby provides a means to assess the LLM's understanding of local connectivity patterns and its ability to evaluate the degree of clustering within a graph. Besides, it tests the ability of reasoning of LLM for computing CC has several steps. 

\begin{figure}[t]
    \centering
    \includegraphics[width=\linewidth]{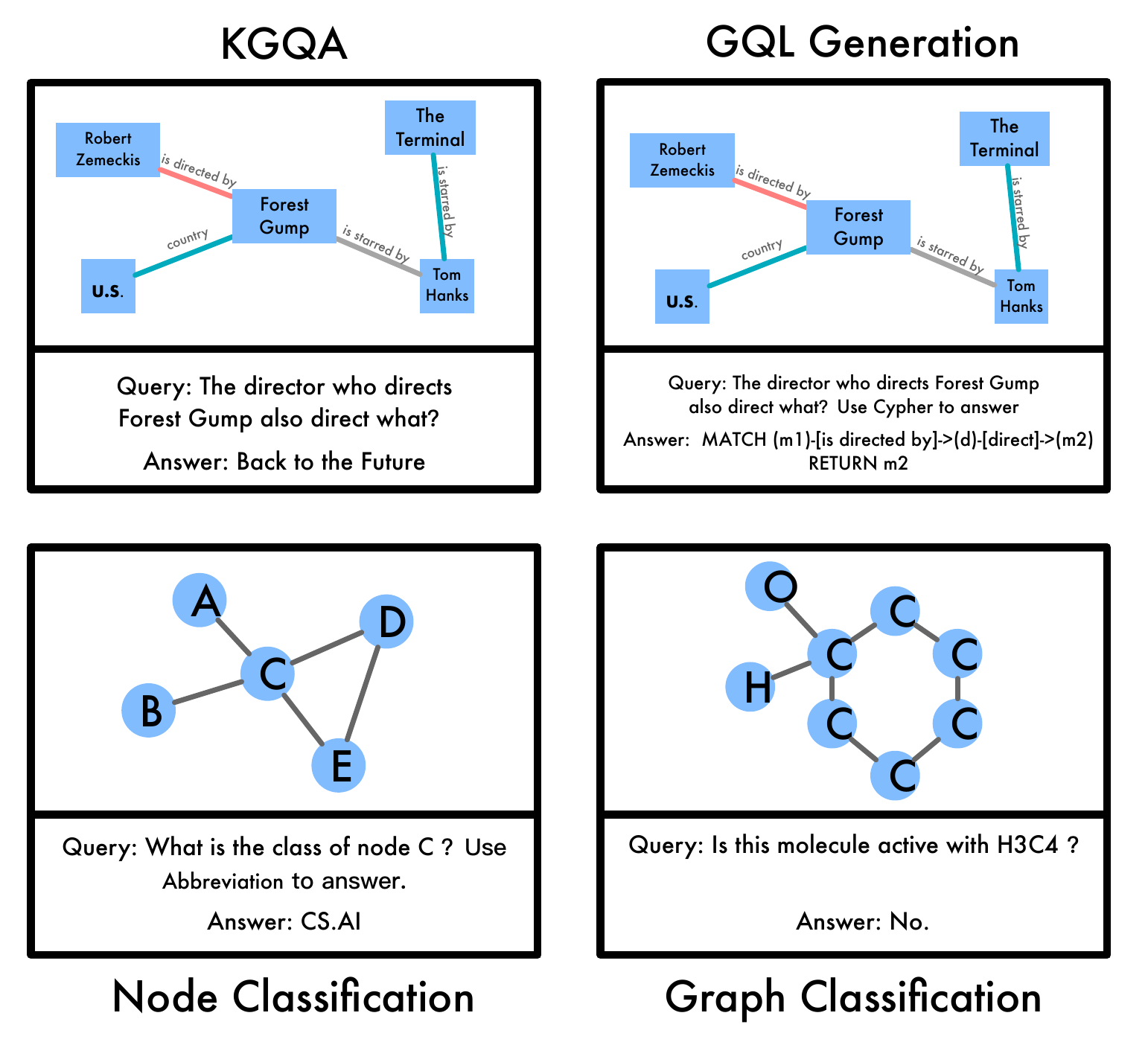}
    \caption{Semantic Understanding Tasks}
    \label{fig:sut}
\end{figure}

\subsection{Semantic Understanding Tasks}

\noindent\textbf{Knowledge Graph Question Answering.} This task gauges the LLM's proficiency in answering questions that pertain to a knowledge graph. Knowledge graphs organize data into a structured format, embodying entities, attributes, and relationships. Task success hinges on the LLM's ability to reason and understand the underlying graph structure to provide accurate answers, thus demonstrating its semantic understanding and capability to navigate and extract information from a KG.

~\

\noindent\textbf{Graph Query Language Generation.} This task measures the LLM's capability to generate graph query languages that satisfy user requirements. These languages, including GQL and Cypher, allow users to extract specific information from a graph database. By generating appropriate graph queries in response to user information needs, the LLM showcases its comprehension of user intent and precision in query formulation.

~\

\noindent\textbf{Node Classification.} This task requires the LLM to classify nodes within a graph based on their attributes or structural features. The LLM is given labeled node examples and their associated classes, and it must correctly predict the class of unseen nodes by applying learned patterns from the labeled data. Success in node classification showcases the LLM's ability to generalize from examples and apply its understanding of node attributes and structure to classify new nodes accurately.

~\

\noindent\textbf{Graph Classification.} This task extends the scope of node classification to encompass entire graphs. The LLM is given graphs, each labeled with specific categories or classes, and is expected to accurately classify unseen graphs by using patterns learned from the labeled examples. This task evaluates the LLM's ability to understand and apply the structural and attribute-based characteristics of a graph holistically, thus enabling accurate classification of new graphs.

\begin{table*}[t]
\Large
    \centering
    \caption{Experiments on Graph Structural Understanding on OGBN-ARXIV. ACC indicates average accuracy over samples, while $\Delta$ indicates the difference of variants with the 1-shot setting. - denotes that the input format do not contain corresponding information.}
    \resizebox{.95\linewidth}{!}{
    \begin{tabular}{cccccccccccccc}
    \toprule
    \multirow{2}{*}{$\textbf{Format}$} & \multirow{2}{*}{$\textbf{Input Design}$} & \multicolumn{2}{c}{$\textbf{Size Detection}$} & \multicolumn{2}{c}{$\textbf{Degree Detection}$} & \multicolumn{2}{c}{$\textbf{Edge Detection}$} & \multicolumn{2}{c}{$\textbf{Attribute Retrieval}$} & \multicolumn{2}{c}{$\textbf{Diameter}$} & \multicolumn{2}{c}{$\textbf{Clustering}$} \\
     & & ACC & $\Delta$ & ACC & $\Delta$ & ACC & $\Delta$ & ACC & $\Delta$ & ACC & $\Delta$ & ACC & $\Delta$ \\
      \midrule
      \multirow{6}{*}{$\textbf{\makecell{Adjacency \\ List}}$} 
      % & 1-shot-sc & \\
       & 1-shot & 35.50 & 0.00 & 15.21 & 0.00 & 65.45 & 0.00 & - & - & 28.00  & 0.00 & 5.42 & 0.00 \\
      \cmidrule{2-14}
      & 1-shot-cot & 44.00 & +8.50 & 14.58 & -0.63 & 65.25 & -0.20 & - & - & 24.00 & -4.00 & 1.85 & -3.57 \\
      & w/o format explanation & 33.00 & -0.25 & 16.34 & +1.13 & 57.50 & -8.25 & - & - & 18.00 & -10.00 & 5.19 & +3.43 \\
     & w/o role prompting & 36.60 & +1.10 & 15.70 & +0.49 & 55.00 & -10.45 & - & - & 20.00 & -8.00 & 4.71 & -0.23 \\
     & w/o change order & 14.00 & -21.50 & 26.28 & +11.07 & 51.20 & -14.25 & - & - & 30.00 & +2.00 & 14.92 & -9.50 \\
     & w/o 1-shot & 33.00 & -2.50 & 17.18 & +1.97 & 71.90 & -6.45 &  - & - & 22.00 & -6.00 & 7.85 & +2.43 \\
    \midrule
     \multirow{6}{*}{$\textbf{Edge List}$} 
     & 1-shot & 22.50 & 0.00 & 44.87 & 0.00 & 74.60 & 0.00 & - & - & 43.00 & 0.00 & 13.31 & 0.00 \\
      \cmidrule{2-14}
      & 1-shot-cot & 27.00 & +4.50 & 48.65 & +3.78 & 74.70 & +0.10 & - & - & 41.00 & -2.00 & 11.33 & -1.98 \\
      & w/o format explanation & 25.00 & +2.50 & 47.86 & +2.99 & 71.55 & -3.05 & - & - & 36.00 & -7.00 & 18.11 & +4.80 \\
     & w/o role prompting & 18.00 & -4.50 & 47.64 & +2.57 & 71.70 & -2.90 & - & - & 39.00 & -4.00 & 13.63 & +0.35 \\
     & w/o change order & 9.00 & -13.50 & 20.48 & -23.39 & 79.60 & +5.00 & - & - & 10.00 & -33.00 & 20.06 & + 7.05 \\
     & w/o 1-shot & 23.00 & +0.50 & 49.34 & +4.47 & 80.95 & +6.35 & - & - & 34.00 & -9.00 & 19.16 & +5.84 \\
    \midrule
    \multirow{6}{*}{$\textbf{GML}$} 
     % & 1-shot-sc & \\
    & 1-shot & 54.50 & 0.00 & 20.91 & 0.00 & 50.45 & 0.00 & 83.40 & 0.00 & 37.00 & 0.00 & 4.36 & 0.00 \\
       \cmidrule{2-14}
       & 1-shot-cot & 55.50 & +1.00 & 20.76 & -0.15 & 50.10 & -0.35 & 83.30 & -0.10 & 28.00 & -9.00 & 0.95 & -3.41 \\
      & w/o format explanation & 55.00 & -0.50 & 29.06 & +8.15 & 50.00 & -0.45 & 85.97 & +2.57 & 41.00 & +4.00 & 12.71 & +8.35 \\
     % & w/o partition mark & \\
     & w/o role prompting & 54.50 & -0.50 & 29.79 & +8.88 & 50.00 & -0.45 & 84.50 & +0.10 & 35.00 & -2.00 & 6.96 & +2.60 \\
     & w/o change order & 51.50 & -3.00 & 21.16 & +0.24 & 55.65 & +5.20 & 83.56 & +0.16 & 39.00 & +2.00 & 5.25 & +0.89 \\
     & w/o 1-shot & 54.00 & -0.50 & 19.85 & -1.06 & 50.25 & +0.20 & 83.22 & -0.18 & 42.00 & +5.00 & 5.39 & +1.03 \\
    \midrule
    \multirow{6}{*}{$\textbf{GraphML}$} 
    & 1-shot & 25.00 & 0.00 & 40.20 & 0.00 & 62.05 & 0.00 & 83.87 & 0.00 & 34.00 & 0.00 & 9.74 & 0.00 \\
       \cmidrule{2-14}
    & 1-shot-cot & 22.50 & -2.50 & 40.02 & -0.18 & 62.30 & +0.25 & 83.75 & -0.12 & 32.00 & -2.00 & 7.29 & -2.45 \\
      & w/o format explanation & 19.00 & -6.00 & 46.90 & +5.88 & 53.75 & -8.40 & 85.37 & +1.50 & 38.00 & +4.00 & 22.75 & +13.01 \\
     & w/o role prompting & 15.50 & -9.50 & 49.89 & +9.87 & 56.10 & -5.95 & 87.63 & +3.76 & 31.00 & -3.00 & 14.52 & +4.78 \\
     & w/o change order & 8.50 & -16.50 & 30.60 & -9.60 & 65.35 & +3.30 & 9.76 & -4.11 & 43.00 & +9.00 & 8.00 & -1.74 \\
     & 0-shot & 24.50 & -0.50 & 39.59 & -0.61 & 73.95 & +11.90 & 82.90 & -0.97 & 30.00 & -4.00 & 14.32 & +4.58  \\
    \bottomrule
    \end{tabular}}
    \label{tab:exp1}
\end{table*}

\section{Data Collection}

\subsection{Structure Understanding Task}
To demonstrate the capabilities of language models in reasoning over Structure Understanding Tasks, we selected two well-known citation networks: obgn-arxiv~\cite{hu2020open} and Aminer~\cite{tang2008arnetminer}. Our approach involved randomly sampling 100 initial seed nodes from each graph and applying a Depth-First Search (DFS) algorithm to sample 2-hop subgraphs centered around these nodes. Each subgraph consisted of approximately 10-20 nodes and 40 edges. To evaluate the performance of the language model, we assigned it the following tasks within these subgraphs: degree detection, attribute retrieval, clustering, size detection, and diameter estimation. For the first three tasks, the model provided results for each individual node in the subgraphs. However, for size detection and diameter estimation, we computed the results for each entire subgraph. Another task we tackled was Edge Detection. Here, we treated each edge in the graph as a positive sample and randomly selected an edge not present in the graph as a negative sample. We then asked the language model to determine whether a given edge belonged to the subgraph or not, based on the information provided by the subgraph. 

\subsection{Semantic Understanding Task}
Shifting our focus to semantic understanding tasks, we conducted knowledge graph question answering using two widely-used datasets: Wiki, a temporal knowledge graph, and MetaQA, a multi-hop movie knowledge base. These datasets served as a testing ground to evaluate the performance of the language model in these domains. For node classification, we leveraged the original labels available in the ogbn-arxiv dataset. We randomly sampled 100 nodes from the test set and tasked the language model with predicting their labels based on information such as the node's title, abstract, and the text information from its k-hop neighbors. In addition, we explored graph query language generation using the MetaQA dataset. We constructed a graph database from this dataset and prompted the language model to generate corresponding graph query languages (GQL) like Cypher. The generated GQL statements were then executed using the Neo4j engine. Through these experiments, we aime to assess the language model's performance in various tasks related to structural and semantic understanding in graph structured data.

\section{Experiments}

\subsection{Experimental Setup}

\noindent\textbf{Downstream Task.} 

\noindent\textbf{Models.} We evaluate the performance of the recent dominant LLM model, InstructGPT-3~\cite{ouyang2022training}, using versions text-davinci-001, text-davinci-002, and text-davinci003. Unless otherwise speciﬁed, we utilize text-davinci-003 in all experiments. The temperature is set to 0.3 to control the variety of the output.

\subsection{Results for Structure Understanding Task}

The results for the Structure Understanding Task are presented in Table~\ref{tab:exp1}, revealing several significant findings:

\noindent\textbf{Input Design Has a Significant Impact on the Final Result.} Our experiments demonstrate that the design of the input plays a crucial role in determining the performance of the model. By carefully considering the arrangement and organization of the input data, we can substantially influence the model's ability to understand the structural aspects of the task at hand. Fine-tuning the input design can lead to improved performance and more accurate structural understanding.

\noindent\textbf{Role Prompting Generally Improves Performance.} Our findings indicate that incorporating role-prompting techniques generally enhances the model's performance in the Structure Understanding Task. By explicitly guiding the model to focus on specific roles or relationships within the graph, we enable it to extract more meaningful insights and make more accurate predictions. Role prompting serves as an effective mechanism for capturing the nuances of the graph's structure and leveraging that information for improved understanding.

\noindent\textbf{Examples Have Impacts on Graph Understanding.} Similar to previous research that suggests the utility of examples in large language models (LLMs), we discovered that examples also have some extend of positive effects in graph understanding scenarios. However, omitting specific examples and relying on zero-shot learning approaches sometimes yielded more powerful results. This phenomenon can be attributed to the rich inherent information present within the graph itself, which allows the model to grasp the complexities of the structure without the need for explicit examples. Examples, in some cases, can introduce noise, biases, or incomplete information, hindering the model's overall understanding.

\noindent\textbf{The Position of External Knowledge Matters.} We investigated the impact of external knowledge, such as questions, statements, and examples, on graph understanding. Comparing the placement of external knowledge before or after the graph input, we observed that positioning external knowledge before the graph generally led to better performance. Placing external knowledge before the graph provides additional context information, enabling the model to better comprehend the specific graph it needs to handle. Conversely, positioning the graph behind external knowledge may hinder the model's ability to effectively utilize the relevant information, potentially degrading performance.

These findings show the importance of thoughtful input design, the potential benefits of role prompting techniques, the limited impact of examples in graph understanding, and the significance of positioning external knowledge for optimal performance. Understanding these factors can guide future research and inform the development of more effective models for structure understanding tasks.

\begin{table*}[t]
    \centering
    \caption{Performance on KGQA and GQL Generation}
    \resizebox{.85\linewidth}{!}{
    \begin{tabular}{c|cccc}
    \toprule
       Method  & Wiki & MetaQA-1hop & MetaQA-2hop & MetaQA-3hop \\
    \midrule
        SOTA & 64.70 & 97.50 & \textbf{98.80} & 94.80 \\
    \midrule
       zero-shot & 9.23 & 24.75 & 6.37 & 9.72 \\
       zero-shot-cot & 8.71 & 18.41 & 12.86 & 21.89 \\
       zero-shot+graph & 56.38 & 91.69 & 46.82 & 19.40 \\
       zero-shot-cot+graph & 55.63 & 86.16 & 47.36 & 19.29 \\
       zero-shot+graph+change-order & 51.35 & 95.20 &  40.48 & 20.17 \\
       zero-shot-cot+graph+change-order & 56.33 & 95.87 & 47.71 & 23.95 \\
    \midrule
    zero-shot Cypher Generation & - & 30.00 & 10.00 & 13.00 \\
    one-shot Cypher Generation & - & $\textbf{99.00}$ & 77.00 & $\textbf{96.00}$ \\
    \bottomrule
    \end{tabular}}
    \label{tab:kgqa}
\end{table*}

\subsection{Results for Semantic Understanding Task}

The resuls for semantic understanding tasks are shown in Figure~\ref{tab:kgqa} and Figure~\ref{tab:nc}. We have the following discoveries:

\noindent\textbf{Resuts for KGQA and GQL generation}. The results for KGQA and GQL generation is shown in Table~\ref{tab:kgqa}. It's noticeable that current SOTA models consistently show higher performance across all datasets, with scores ranging from 94.80 on MetaQA-3hop to 98.80 on MetaQA-2hop. However, LLM showed comparative performance on certain tasks with prompt strategies. Specifically, the 'zero-shot+graph' method has performed exceptionally well on the 'Wiki' dataset, achieving an accuracy of 56.38, the highest among our proposed models. Similarly, the 'zero-shot-cot+graph+change-order' model performs the best on MetaQA-1hop, scoring 95.87. When we compare zero-shot models with 'zero-shot-cot' counterparts, we observe a general trend that the inclusion of the graph ('+graph') and change order ('+change-order') enhancements improve the model performance. For the 'one-shot Cypher' method, an impressive performance of 99.00 is achieved on the MetaQA-1hop, surpassing the state-of-the-art and all other models in our study. 

~\

\noindent\textbf{Results for Node Classification.} For Node Classification on OGBN-ARXIV (Table~\ref{tab:nc}), the 'one-shot + 1-hop neighborhood context summarization'  model has the highest accuracy of 60.00 among all the variants. Interestingly, models augmented with 2-hop neighborhood context summarization ('2-hop') show better performance than their 1-hop counterparts, showing that expanding context range is helpful in providing valuable information. Also, the  model performs better than the change-of-thought~(cot) model, suggesting that the cot strategy might not be as effective for this task. These results indicate potential areas for improvement, particularly for the 'zero-shot-cot' and 'change-order' strategies, which don't consistently improve performance. Nonetheless, the experiments provide valuable insights into the performance of different strategies in the node classification task.

\begin{table}[t]
    \centering
    \caption{Performance of Node Classification on OGBN-ARXIV. self denotes only the use of the text feature of the target nodes. 1-hop denotes using the text feature of direct neighbors. 2-hop denotes using the text feature within 2-hop neighbors.  }
    \begin{tabular}{ccc}
    \toprule
       Method & Context & ACC \\
    \midrule
       \multirow{3}{*}{zero-shot} & self & 48.00 \\
        &  1-hop & 53.00 \\
        & 2-hop & 57.00 \\
    \midrule
       \multirow{3}{*}{zero-shot-cot} & self & 40.00 \\
        & 1-hop & 40.00 \\
        & 2-hop & 56.00 \\
    \midrule
        \multirow{3}{*}{one-shot} & self & 50.00 \\
        & 1-hop & 54.00 \\
        & 2-hop & 60.00 \\
    \midrule
        \multirow{3}{*}{one-shot-cot} & self & 43.00 \\
        & 1-hop & 55.00 \\
        & 2-hop & 59.00 \\
    \bottomrule
    \end{tabular}
    \label{tab:nc}
\end{table}

\begin{table*}[t]
    \centering
    \caption{Performance on Graph Classification}
    \begin{tabular}{c|cc|cc}
    \toprule
       \multirow{2}{*}{Dataset}  & \multicolumn{2}{c}{OGBG-MOLHIV} &  \multicolumn{2}{c}{OGBG-MOLPCBA} \\
       & GML & GraphML & GML & GraphML \\
    \midrule
       1-shot-tot & 66.87 & 63.25 & 57.18 & 57.45 \\
       1-shot-cot & 67.65 & 64.71 & 59.26 & 57.32\\
    \midrule
        w/o self-format explanation  & 64.71 & 64.71 & 58.73 & 56.24 \\
        w/o self-summarization & 61.76 & 61.77 & 57.64 & 56.67\\
        0-shot-cot & 58.82 & 59.76 & 55.57 & 55.32 \\
    \bottomrule
    \end{tabular}
    \label{tab:gc}
\end{table*}

~\

\noindent\textbf{Results for Graph Classification.} The results for graph classification task is shown in Table~\ref{tab:gc}. From the result, we find that the self-augmentation is effective in improving the performance of GC. It shows that self-augmentation like self-format explanation and self-summarization can enrich the context of the original graph and will make the LLM more easily complete the task. 

\section{Discussion}

 Our findings suggest several promising directions for future work in structure understanding tasks with LLMs. First, more research is needed to understand how different input designs and role prompting techniques can further enhance performance. Second, we encourage researchers to investigate why examples are less effective for graph understanding and to explore alternative strategies for leveraging the rich information embedded in graphs. Third, the role of external knowledge placement merits further exploration. Finally, new approaches for graph augmentation could be developed to improve performance on semantic understanding tasks.

In addition, our experiments have revealed the potential of LLMs in various tasks beyond pure natural language processing. We believe that more effort should be dedicated to integrating graph-based information into LLMs, exploring different types of graph structures, and applying LLMs to other areas such as graph theory, network science, and complex systems. In the future, we may also consider using LLM to control the use of external tools to better handle graph structured data~\cite{schick2023toolformer,zhang2023graph}.

% \section{Insights and Analysis}

% \textbf{Can LLM become a powerful tool in assisting graph analysis?} 

% \textbf{Better ways to utilize LLM for graph understanding. } 

\section{Related Works}

\subsection{Language Model for Structural Data Understanding}

Language models are being extended to understand and work with structural data, such as graphs, tables, and trees. One approach is using graph neural networks to encode structural information, capturing dependencies and relationships between elements~\cite{qasim2019rethinking}. Incorporating GNNs into language models enables them to generate contextually aware outputs that consider the structural characteristics of the data. Another approach is incorporating attention mechanisms into language models for structural data~\cite{chen-etal-2022-towards-table,eisenschlos2021mate}. Attention allows the model to focus on relevant parts, improving understanding of complex dependencies and enhancing performance in tasks like graph completion and table understanding. Language models can also benefit from combining knowledge graph embeddings with textual information, leveraging both textual and structural data to make informed predictions.

\subsection{Graph Machine Learning}

Graph machine learning develops models and algorithms for data structured as graphs, representing complex relationships in various domains. Traditional machine learning struggles with graph-structured data, but graph machine learning methods utilize the graph structure to extract meaningful features and make predictions. Graph convolutional~(GCN) networks extend convolutional neural networks to operate on graph-structured data, capturing local and global structural patterns and excelling in tasks like node classification and graph-level classification~\cite{kipfsemi}. Graph attention networks (GAT) incorporate attention mechanisms, allowing adaptive aggregation of information from relevant nodes~\cite{velickovic2017graph}. GAT perform well in tasks like node classification and graph-level representation learning. Graph generative models generate new graphs to capture the structural characteristics and properties of the input data, benefiting tasks like molecule generation~\cite{walters2020applications} and graph-based data augmentation~\cite{zhao2021data}. Graph machine learning techniques enable effective analysis and extraction of insights from graph-structured data, advancing fields relying on understanding complex relationships and dependencies.

\section{Conclusion}

In this work, we analyze the ability of large language models to understand graph-structured data. Our findings indicate that there is still a long way for a LLM to understand graph data. Future research should focus on developing and refining methods for encoding graph-structured information into a format that a large language model can comprehend and manipulate effectively. This is a complex challenge given the inherent differences between sequential text data and graph data, which is intrinsically multi-dimensional and relational.

\bibliographystyle{acl_natbib}
\bibliography{custom}

\appendix
\section{Detailed Description of Datasets}

\subsection{OGBN-ARXIV}

The Open Graph Benchmark (OGB) is a collection of diverse, large-scale, and challenging datasets and benchmarking tasks for graph machine learning research. OGBN-ARXIV is a part of the OGB Node Property Prediction track. The dataset comprises academic papers from the arXiv website, which are represented as nodes in a citation graph. In the graph, the edges denote the citation relationships between the papers. Each paper is associated with a 128-dimensional word2vec feature vector derived from its title and abstract. The task associated with this dataset is to predict the subject area of each paper, making it a multi-class classification problem. We sample a subset of 100 nodes with multi-hop neighbors for testing.

\subsection{OGBG-MOLX}

OGBG-MOLX is part of the Graph Property Prediction track in OGB and it comprises of two datasets: MOLHIV and MOLPCBA. MOLHIV dataset contains molecular graphs where the task is to predict whether a molecule inhibits HIV virus replication or not, making it a binary classification problem. On the other hand, MOLPCBA dataset contains molecular graphs with the task of predicting bioactivity for various protein targets, which is a multi-label classification problem. In these datasets, nodes represent atoms and edges represent bonds between atoms. Node and edge features include atom type, atom degree, bond type, and whether the bond is in a ring. We sample 100 graphs with the same number of positive and negative samples for testing.

\subsection{Wiki}

The Wiki dataset is a well-known dataset that contains text from a collection of Wikipedia articles. The structure of this dataset varies depending on the particular task. For instance, for text classification tasks, each document (or article) can be represented as a bag-of-words vector, with each dimension representing the frequency of a specific word. The labels may include the categories that the articles belong to. In the context of graph-based tasks, a Wikipedia dataset could consist of a network of articles (as nodes) linked by hyperlinks (as edges), with the task being predicting article categories or link prediction between articles.

\subsection{MetaQA}

MetaQA is a dataset designed for the task of multi-hop reasoning in question answering. It consists of movie-related knowledge graph entities, relations, and natural language questions. Each node in the knowledge graph represents a movie entity (such as a movie, actor, or director), and edges represent relationships between entities. The dataset also includes questions at three levels of complexity (1-hop, 2-hop, and 3-hop), with each level requiring reasoning over an increasing number of edges in the graph to answer the questions correctly. The goal is to answer natural language questions by effectively reasoning over the underlying knowledge graph.

\begin{table}[t]
\small
\caption{Input Design for Different Tasks.}
    \centering
    \begin{tabularx}{\linewidth}{lXX}
    \toprule
    \textbf{Task} & \textbf{Description} \\
    \midrule
       Size Detection & <graph text> What is the number of nodes and edges in this graph? Please answer with the number of nodes: X, number of edges: X. \\
    \midrule
       Degree Detection  & <graph text> What is the degree of node X? \\
    \midrule
        Edge Detection & <graph text> Is there an edge between node X1 and node X2? \\
    \midrule
        Attribute Retrieval & What is the title of node X? \\
    \midrule
        Diameter & What is the diameter of this graph? \\
    \midrule
        Clustering & What is the clustering coefficient of node X? \\
    \midrule
        KGQA & Knowledge: <graph text>, Question: <question text> \\
    \midrule
        GQL Generation & Thus the Neo4j CQL of the question is \\
    \midrule
        Node Classification & Which arxiv CS subcategory does paper <paper title> with abstract <paper abstract> belongs to? use the abbreviation to answer. \\
    \midrule
        Graph Classification & <graph text> Whether the molecule inhibits HIV virus replication? Yes or no. \\
    \bottomrule
    \end{tabularx}
    \label{tab:description}
\end{table}

\section{Input Design for Different Tasks}

The input design for different tasks are shown in Table~\ref{tab:description}, where we show the question designs for different tasks.

\section{Cypher Introduction}

Cypher is a declarative graph query language developed by Neo4j, a popular graph database management system. It allows for expressive and efficient querying and updating of graph data. The language is designed to be intuitive and readable, drawing on the use of English prose and iconography. Cypher is built around the concept of pattern matching. It focuses on the clarity of expressing what to retrieve from a graph, not dictating how to retrieve it. This design makes Cypher powerful when working with graph data, as patterns are often more intuitive and easier to understand. 

\section{Data and Code Release}

The GUC benchmark and codes in this paper will be open sourced at~\url{https://anonymous.4open.science/r/GPT4Graph}. after an internal review. The synthesized labels in the benchmark will be released under CDLAPermissive-2.0 license. Our code will be released publicly with MIT license.

% \section{Case Study}

% \subsection{Clustering Coefficient Computing}

% \subsection{}

\end{document}